%% file: root.tex
\documentclass[letterpaper,10pt,conference]{ieeeconf}

\IEEEoverridecommandlockouts
\usepackage{graphicx}
\usepackage{subfig}
\usepackage{booktabs}
\usepackage{amsmath}
\usepackage{amssymb}
\usepackage{placeins}
\usepackage{flafter}
\usepackage{stfloats}
\usepackage[hidelinks]{hyperref}
\hypersetup{
  pdftitle={Beyond Safe Execution: Safe Online Reinforcement Learning for Autonomous Driving with Value Calibration and Adaptive Guidance},
  pdfauthor={Zeqiao Li, Yijing Wang, Haoyu Wang, Zheng Li, Zhiqiang Zuo},
  pdfsubject={Safe online reinforcement learning for autonomous driving},
  pdfkeywords={SOAR-Flow, reinforcement learning, autonomous driving, safety filtering, value calibration}
}

\newtheorem{theorem}{Theorem}

\title{\LARGE \bf Beyond Safe Execution: Safe Online Reinforcement Learning
for Autonomous Driving with Value Calibration and Adaptive Guidance}
\author{Zeqiao Li, Yijing Wang, Haoyu Wang, Zheng Li, and Zhiqiang Zuo}

\begin{document}

\maketitle
\thispagestyle{plain}
\pagestyle{plain}

\begin{abstract}
Online reinforcement learning (RL) enables driving policies to adapt
to changing deployment conditions, but requires safe exploration
throughout training. Safety filtering alone does not ensure stable
policy improvement or reduced dependence on corrections. We propose
Safe Online Adaptive Refinement (SOAR-Flow), a framework for safe
and stable online fine-tuning of offline-initialized Flow-based
driving policies. A control barrier function (CBF)--model predictive
control (MPC) filter corrects potentially unsafe actions before
execution. To support stable policy updates, we introduce
Gaussian-return-aware value calibration based on calibrated
Q-learning (Cal-QL), using recorded behavior returns and predicted
return variability to smooth conservative critic updates and mitigate
policy degradation. Paired corrective learning teaches feasible action
adjustments, while an intervention-to-go predictor encourages decisions
requiring fewer future corrections. Its predictions also regulate
guidance strength, allowing recovery supervision to weaken as predicted
dependence decreases. Together with gradually reduced calibration
regularization, this supports continued task improvement while
preserving hard execution constraints. We analyze conditions for
safe execution at control steps and diminishing auxiliary learning
influences. We evaluate SOAR-Flow in simulation and on an uncrewed
ground vehicle (UGV). The results demonstrate improved driving
performance and reduced reliance on safety corrections during
protected online fine-tuning.
The demonstration video and project overview are available at
\url{https://lzqw.github.io/SOAR-Flow/}.
\end{abstract}

\input{sections/introduction}
\input{sections/related_work}
\input{sections/methodology}

\input{sections/simulation_experiments}

\input{sections/ugv_experiments}
\FloatBarrier
\input{sections/conclusion}

\FloatBarrier
\bibliographystyle{IEEEtran}
\bibliography{references}

\end{document}

%% file: sections/introduction.tex
\section{Introduction}
\label{sec:introduction}

Learning safe and efficient driving behavior in changing environments
remains a central challenge in autonomous driving~\cite{kiran2022survey}.
Online RL enables continued policy improvement through interaction,
with offline demonstrations providing useful initial
skills~\cite{calql,flow_q_learning}. On UGVs, however,
trial-and-error exploration can lead to collisions and other accidents,
making safety essential throughout online learning~\cite{h_dsac}.

Preventing failures before they occur is a practical approach to safe
exploration. Failure-Aware Offline-to-Online Reinforcement Learning
(FARL) follows this principle through failure prediction and
preventive recovery in robotic manipulation~\cite{farl2026}. Model-based
safety filters check actions against explicit constraints and compute
feasible corrections before execution~\cite{safety_filter_training}.
However, successful protected execution can still depend on repeated
corrections, while online updates may degrade previously learned
behavior~\cite{calql,safety_filter_training}. Going beyond safe execution
therefore requires learning effectively from protected experience to
improve task performance and reduce correction demand.

Reliable value estimates are central to this improvement. Overestimation
may favor actions with limited data support, whereas excessive pessimism
can undervalue established behaviors as new experience is
incorporated~\cite{calql}. Filtering also changes the executed action,
so observed rewards and transitions need to be associated with the action
actually applied~\cite{safety_filter_training}. Cal-QL addresses value-scale
mismatch by calibrating conservative estimates against a reference
behavior's return~\cite{calql}. This motivates calibration that uses
recorded execution outcomes while accounting for return variability
during continued online learning.

Learning from corrected actions can reduce repeated
interventions~\cite{safety_filter_training}. Human--AI Copilot Optimization
(HACO) estimates cumulative human intervention costs to guide policy
optimization~\cite{haco}, motivating consideration of future assistance
alongside immediate corrections. In driving, earlier position or speed
choices can determine whether later corrections are needed. Moreover,
optimizing a finite-horizon MPC cost need not maximize long-horizon task
return. Corrective guidance should therefore
combine feasible local adjustments with anticipation of future
interventions, while allowing its influence to diminish as dependence
decreases. This leaves room for continued reward-driven improvement
under unchanged safety constraints.

To address these challenges, we propose SOAR-Flow, a framework for safe
online refinement of offline-initialized Flow-based driving policies.
CBF--MPC certifies references before execution to protect online
interaction. We introduce Gaussian-return-aware value calibration by
extending Cal-QL with the Gaussian return model of distributional soft
actor--critic (DSAC)~\cite{calql,dsac}. Recorded behavior returns provide
calibration references, while predicted return variability smooths
conservative critic updates. Paired corrective learning teaches feasible
action adjustments, and an intervention-to-go predictor encourages earlier
adjustments to reduce subsequent corrections. The same predictions
regulate guidance strength, allowing recovery supervision to weaken as
predicted dependence decreases. Together with decaying calibration
regularization, this supports continued task improvement under unchanged
execution constraints. The main contributions are as follows:
\begin{itemize}
    \item A Gaussian-return-aware extension of Cal-QL combines recorded
    behavior returns with predicted return variability to smooth
    conservative critic updates and support stable online learning from
    protected execution.

    \item An adaptive guidance mechanism combines paired corrective
    learning with intervention-to-go prediction. The predictor guides
    policy actions and regulates guidance strength to reduce filter
    reliance while retaining task optimization.

    \item We establish conditions for sample-time safety of protected
    execution and vanishing auxiliary gradient contributions.
    Safety-filtered (SF) and policy-only (PO) tests in MetaDrive and on a UGV
    assess learning stability, safety, and autonomy with nominal MPC
    tracking retained.
\end{itemize}

%% file: sections/related_work.tex
\section{Related Work}
\label{sec:related_work}

\textbf{Safe RL and intervention-based learning.}
Safe RL addresses safety through policy optimization and execution-time
protection. Constrained policy optimization regulates expected costs
in policy updates~\cite{cpo}, whereas recovery policies and predictive
safety filters intervene using learned risk estimates or model-based
feasibility checks~\cite{recovery_rl,predictive_safety_filter}.
For pretrained policies, SafeFlow enforces barrier conditions during
test-time flow-based trajectory generation~\cite{dai2025safeflow},
while FARL uses an offline-trained world model and recovery policy
to prevent failures during online refinement, keeping both protection
components fixed online~\cite{farl2026}. Beyond execution protection,
interventions also provide information for policy learning.
Safety Filtering While Training studies how filtering and correction
penalties affect learned behavior~\cite{safety_filter_training}, while
CBF-RL combines training-time filtering with barrier-inspired rewards
to encourage safer actions without runtime correction~\cite{cbf_rl}.
HACO estimates cumulative human intervention costs to guide policy
optimization~\cite{haco}. Building on this connection between protection
and learning, SOAR-Flow combines paired corrective supervision with
prediction of future filter use. The predictions guide policy actions
and regulate supervision strength, while CBF--MPC certification
remains active.

\textbf{Value learning and online policy refinement.}
Online refinement seeks to improve from new experience while retaining
useful pretrained behavior. Offline RL addresses unsupported actions
through conservative value estimation in conservative Q-learning or
in-dataset value learning in implicit Q-learning~\cite{kumar2020cql,iql}.
However, excessive pessimism can hinder the transition to online learning,
motivating Cal-QL's calibration against a reference behavior's
return~\cite{calql}. Policy regularization and objective alignment offer
complementary approaches: PROTO evolves its policy regularizer during
fine-tuning~\cite{proto}, while Uni-O4 uses a shared on-policy objective
across offline and online stages~\cite{uni_o4}. Generative policies
complement these optimization methods by representing diverse behaviors
in prior data. Flow Q-learning combines flow-based action modeling with
value-driven training of an expressive one-step actor~\cite{flow_q_learning};
RL-100 connects imitation, iterative offline RL, and online refinement
for diffusion-based robotic policies~\cite{rl100}. Building on Cal-QL
and DSAC's Gaussian return model~\cite{calql,dsac}, SOAR-Flow uses
predicted return variability to smooth behavior-referenced calibration
of values learned from certified execution. This value-learning objective
complements adaptive intervention guidance during protected online
policy refinement.

%% file: sections/methodology.tex
\section{Methodology}
\label{sec:soar_flow}

Building on Cal-QL~\cite{calql}, SOAR-Flow performs safe online refinement
of a Flow policy and task critics initialized from offline data.
CBF--MPC checks policy actions before execution. The resulting
transitions support Gaussian-return-aware value calibration, while
corrections and intervention sequences guide policy improvement.
Adaptive guidance reduces auxiliary learning influences as predicted
filter dependence decreases. Fig.~\ref{fig:framework} summarizes this
protected interaction and update loop.

\input{sections/framework_placeholder}

Autonomous driving is modeled as a continuous-action Markov decision process (MDP)
$\mathcal M=\langle\mathcal S,\mathcal A,\mathcal P,\mathcal R,\gamma\rangle$,
where $\mathcal S$ and $\mathcal A$ are the state and driving-reference
spaces, $\mathcal P$ is the transition kernel, $\mathcal R$ is the task
reward function, and $\gamma\in[0,1)$ is the discount factor.
The policy maximizes the entropy-augmented objective of soft actor--critic
(SAC)~\cite{SAC}
\begin{equation}
    J_\pi=\mathbb E_{(s_{i\ge t},a_{i\ge t})\sim\rho_\pi}
    \left[\sum_{i=t}^{\infty}\gamma^{i-t}
    \left[r_i+\alpha\mathcal H\bigl(\pi(\cdot\mid s_i)\bigr)\right]\right],
    \label{eq:task_return_objective}
\end{equation}
where $r_i$ is the task reward, $\rho_\pi$ is the policy-induced
trajectory distribution, $\alpha\ge0$ is the temperature coefficient,
and $\mathcal H(\pi(\cdot\mid s))=
\mathbb E_{a\sim\pi(\cdot\mid s)}[-\log\pi(a\mid s)]$.
The optimal policy satisfies $\pi^\star\in\arg\max_\pi J_\pi$.

A conditional Flow policy transforms Gaussian noise into driving
references through a state-conditioned vector field
$v_\phi$~\cite{flow_q_learning,flow_matching}. Given state $s$ and
$z\sim\mathcal N(0,I)$, the flow satisfies
\begin{equation}
    \frac{\mathrm d x_\tau}{\mathrm d\tau}
    =v_\phi(x_\tau,\tau,s),
    \qquad x_0=z,\quad \tau\in[0,1],
    \label{eq:conditional_flow}
\end{equation}
where $\tau$ denotes flow time. The endpoint
$\mu_\phi(s,z)=x_1$ induces the policy distribution
$\pi_\phi(\cdot\mid s)$.
Given an offline dataset $\mathcal D_{\mathrm{off}}$ of state--reference
pairs, we sample $(s,a)\sim\mathcal D_{\mathrm{off}}$ and
$\tau\sim\mathcal U[0,1]$ independently of $z$, and let
$\widetilde x_\tau=(1-\tau)z+\tau a$. Flow-based behavior cloning
(Flow-BC) initializes the policy by minimizing
\begin{equation}
    \mathcal L_{\mathrm{FM}}(\phi)
    =\mathbb E\!\left[
        \left\|v_\phi(\widetilde x_\tau,\tau,s)-(a-z)\right\|_2^2
    \right].
    \label{eq:flow_matching_loss}
\end{equation}

\subsection{Calibrated Value Learning under Safety Filtering}
\label{subsec:calibrated_learning}
\textbf{Maintaining safe exploration.}
CBF--MPC checks policy actions using a prescribed nominal model and
the same MPC tracking solution used for execution~\cite{safety_filter_training}.
Let $\xi_t$ denote the prediction state. Barrier functions $h_j$ define
$\mathcal C=\{\xi\in\mathcal X:h_j(\xi)\ge0,\ \forall j\}$,
where $\mathcal X$ is the admissible state set. For reference $a$,
the predicted tracking trajectory satisfies the discrete-time CBF
condition~\cite{cbf_rl,discrete_cbf}
\begin{equation}
    h_j(\hat\xi^a_{k+1\mid t})
    \ge\rho_jh_j(\hat\xi^a_{k\mid t}),
    \qquad \rho_j\in[0,1).
    \label{eq:predicted_cbf_residual}
\end{equation}

Here, $\hat\xi^a_{0\mid t}=\xi_t$, $k=0,\ldots,N_p-1$, and
$j=1,\ldots,n_h$, with prediction horizon $N_p$.
The set $\mathcal A_{\mathrm{cert}}(\xi_t)$ contains references whose
tracking solutions satisfy this condition, admissible state and input
constraints, and a terminal condition in $\mathcal X_f\subseteq\mathcal C$.
The terminal set admits a feasible backup controller that keeps the
nominal state within it. For example, if tracking a reference would
violate obstacle clearance, the filter searches for a feasible bypass
reference before execution. Online interaction follows
\begin{equation}
\begin{aligned}
    a_t^{\mathrm{raw}}
    &\sim\pi_\phi(\cdot\mid s_t),\\
    a_t^{\mathrm{cert}}
    &=\mathcal F_{\xi_t}(a_t^{\mathrm{raw}}),
\end{aligned}
\label{eq:protected_interaction}
\end{equation}
where $\mathcal F_{\xi_t}$ retains certifiable references or returns
feasible corrections. The first input of the associated MPC tracking
solution is applied as $u_t=u^\star_{0\mid t}$, with certification
repeated at the next control step. We store
$\mathcal T_t=(s_t,a_t^{\mathrm{raw}},a_t^{\mathrm{cert}},
\iota_t,r_t,s_{t+1},d_t)$, where $\iota_t=1$ denotes a
certification-triggered correction and $d_t=1$ denotes true termination.
The generator input and prediction states are retained for paired
updates and target certification.
These transitions form $\mathcal D_{\mathrm{on}}$, with mixed replay
$\mathcal D=\mathcal D_{\mathrm{off}}\cup\mathcal D_{\mathrm{on}}$.
The reward $r_t$ and next state $s_{t+1}$ result from tracking
$a_t^{\mathrm{cert}}$ and therefore supervise the task critics at
$(s_t,a_t^{\mathrm{cert}})$~\cite{safety_filter_training}.
Recorded corrections provide action targets for learning references
that pass certification directly.

\textbf{Avoiding policy degradation.}
Cal-QL uses a reference behavior's return to limit excessive pessimism
during O2O learning~\cite{calql}. For each transition
from a completed episode ending at $T$, we store the discounted task
return $b_t=\sum_{k=t}^{T-1}\gamma^{k-t}r_k$ as its calibration
reference. This quantity is computed from recorded offline and online
trajectories; for online data, it includes rewards from both unmodified
and corrected execution. Cal-QL's candidate term
$\max\{Q(s_t,a),b_t\}$ retains conservative pressure above the
reference and stops it below, producing an abrupt change in its mean
gradient. We extend this construction with DSAC's Gaussian return
model~\cite{dsac} to smooth calibration using predicted return
variability. For return variable $Z^\pi(s,a)$, its mean satisfies
$Q^\pi(s,a)=\mathbb E[Z^\pi(s,a)]$. Each critic predicts the
mean $Q$ and variance $\sigma^2>0$ of a Gaussian approximation:
\begin{equation}
    \mathcal Z(\cdot\mid s,a)
    =\mathcal N\!\left(Q(s,a),\sigma^2(s,a)\right).
    \label{eq:gaussian_task_critics}
\end{equation}

For a candidate $a\sim\pi_\phi(\cdot\mid s_t)$,
let $Q_i=Q_{\theta_i}(s_t,a)$ and
$\sigma_i=\sigma_{\theta_i}(s_t,a)>0$ denote the outputs of
critic $i\in\{1,2\}$. We define
\begin{equation}
\begin{aligned}
    \mathcal G_i(s_t,a;b_t)
    &:={\mathbb E}_{\epsilon\sim\mathcal N(0,1)}
       [\max\{Q_i+\sigma_i\epsilon,b_t\}]\\
    &=b_t+(Q_i-b_t)\Phi(\nu_i)+\sigma_i\varphi(\nu_i).
\end{aligned}
\label{eq:gaussian_calibration_function}
\end{equation}

The standardized offset is $\nu_i=(Q_i-b_t)/\sigma_i$; $\Phi$ and
$\varphi$ denote the standard normal cumulative distribution and density functions.
For fixed reference and standard deviation (SD),
$\partial\mathcal G_i/\partial Q_i=\Phi(\nu_i)$, so the
conservative pressure changes continuously around $b_t$.
The predicted SD determines the smoothing width,
and the mean-only truncation is recovered in the zero-variance limit.
Gradients through $\sigma_i$ are disabled in $\mathcal G_i$ to
avoid directly shrinking the return dispersion; the SD
remains trainable in the Bellman term.

At the recorded execution pair, write
$Q_i^{\mathrm{cert}}=Q_{\theta_i}(s_t,a_t^{\mathrm{cert}})$ and
$\sigma_i^{\mathrm{cert}}=\sigma_{\theta_i}(s_t,a_t^{\mathrm{cert}})$.
The complete critic objective is
\begin{equation}
\begin{aligned}
    \mathcal L_{\mathrm{task},i}
    ={}&\beta_{\mathrm{cal},n}\mathbb E_{\mathcal D,a}
    \!\left[\mathcal G_i(s_t,a;b_t)-Q_i^{\mathrm{cert}}\right]\\
    &+\mathbb E_{\mathcal D,a',Z'}\!\left[
        \frac{(Y_t^Z-Q_i^{\mathrm{cert}})^2}
             {2(\sigma_i^{\mathrm{cert}})^2}
        +\log\sigma_i^{\mathrm{cert}}
    \right],
\end{aligned}
\label{eq:calibrated_task_critic}
\end{equation}
where $\beta_{\mathrm{cal},n}\ge0$ controls calibration at update $n$.
The first expectation samples current policy actions at replay states
and uses only records with available $b_t$; its unweighted value is
$\mathcal R_{\mathrm{GCal},i}$. The second uses all replay transitions
to fit the Gaussian return distribution. Its mean gradient is scaled
by the inverse predicted variance. Together, these terms fit recorded
execution outcomes while calibrating candidate values against observed
behavior.

The Bellman target follows the same protected execution rule as data
collection. For $(s',\xi')=(s_{t+1},\xi_{t+1})$, we use
\begin{equation}
\begin{aligned}
    a'&\sim\pi_{\bar\phi}(\cdot\mid s'),
    \quad a'^{\mathrm{cert}}=\mathcal F_{\xi'}(a'),\\
    Z'&\sim\overline{\mathcal Z}
        (\cdot\mid s',a'^{\mathrm{cert}}),\\
    Y_t^Z&=r_t+\gamma(1-d_t)Z'.
\end{aligned}
\label{eq:protected_distributional_target}
\end{equation}

Bars denote target quantities. At each
$(s',a'^{\mathrm{cert}})$, $\overline{\mathcal Z}$ is the
Gaussian distribution of the target critic with the smaller mean,
using that same critic's variance~\cite{dsac_t}.
The sample $Z'$ represents the continuation return under protected
execution; terminal transitions use $Y_t^Z=r_t$.
The objective is used for both offline initialization and online
refinement, with actor parameters, Bellman targets, and recorded
behavior returns fixed during critic updates.

\subsection{Adaptive Guidance for Autonomous Policy Improvement}
\label{subsec:adaptive_improvement}

Calibrated critics support policy improvement from protected experience,
but successful execution can still rely on repeated corrections.
Moreover, minimizing a finite-horizon MPC cost need not maximize
long-horizon task return.
We therefore use corrections as local guidance while allowing the policy
to discover higher-return behavior within the prescribed safety constraints.

\textbf{Learning from task values and corrections.}
Online records provide two complementary learning signals. For
$\iota=0$, the task critic guides improvement of the policy action;
for $\iota=1$, the executed correction provides a feasible action target.
Including the entropy term in \eqref{eq:task_return_objective}, we use
\begin{align}
    \mathcal L_Q(\phi)
    ={}&-\mathbb E_{\mathcal D_{\mathrm{on}}}
    \!\left[\bar Q_\theta(s,\mu_\phi(s,z))\mid\iota=0\right]
    \nonumber\\
    &-\alpha\mathbb E_{s\sim\mathcal D_{\mathrm{on}}}
    \!\left[\mathcal H(\pi_\phi(\cdot\mid s))\right],
    \label{eq:task_actor_objective}\\
    \mathcal L_{\mathrm{corr}}(\phi)
    ={}&\mathbb E_{\mathcal D_{\mathrm{on}}}
    \!\left[\|\mu_\phi(s,z)-a^{\mathrm{cert}}\|_{W_a}^{2}
    \mid\iota=1\right].
    \label{eq:intervention_corrective_loss}
\end{align}

The clipped value is $\bar Q_\theta=\min_iQ_{\theta_i}$, $\alpha$ is the
entropy coefficient, and $W_a\succ0$ weights the corrective error.
The recorded generator input $z$ is reused for
these paired updates, associating each correction with the action
that triggered it. Entropy is evaluated using fresh samples from the
current raw policy. The intervention labels select replay records;
updated actions are checked again during execution.

\textbf{Anticipating future interventions.}
An earlier position or speed choice can determine whether a later
correction is needed. Motivated by intervention-value
learning~\cite{haco}, we estimate discounted future filter use:
\begin{equation}
\begin{aligned}
    &P^\pi(s_t,a_t^{\mathrm{raw}})\\
    &\quad=(1-\gamma_P)\mathbb E_{\pi,\mathcal F}\!\left[
    \sum_{\ell=0}^{\infty}\gamma_P^\ell\iota_{t+\ell}
    \,\middle|\,s_t,a_t^{\mathrm{raw}}\right].
\end{aligned}
\label{eq:filter_dependence_value}
\end{equation}

The discount factor satisfies $\gamma_P\in[0,1)$; subsequent actions follow $\pi$, and
transitions follow certified execution. Interventions after termination
are zero, giving $P^\pi\in[0,1]$.
To propagate delayed interventions, we use contiguous segments of
$K$ transitions ending at termination or a collection boundary.
The observed interventions provide a partial target, and the endpoint
estimate supplies the remaining continuation:
\begin{equation}
\begin{aligned}
    y_t^P={}&(1-\gamma_P)\sum_{\ell=0}^{K-1}
    \gamma_P^\ell\iota_{t+\ell}\\
    &+\gamma_P^K(1-d_{t+K-1})
    \mathbb E_{a'}[P_\psi(s_{t+K},a')],\\
    \mathcal L_P(\psi)={}&\tfrac12
    \mathbb E_{\mathcal D_{\mathrm{on}}}
    \!\left[(P_\psi(s_t,a_t^{\mathrm{raw}})-y_t^P)^2\right].
\end{aligned}
\label{eq:dependence_learning}
\end{equation}

The bootstrap action $a'\sim\pi_\phi(\cdot\mid s_{t+K})$ is sampled from the raw policy.
True termination removes the continuation term. Targets are fixed during
regression, and $P_\psi$ is bounded in $[0,1]$.
Setting $K=1$ gives the one-step target; longer segments propagate
observed intervention signals farther in each update.
These targets approximate current-policy evaluation from collected data.

The complete actor objective is
\begin{equation}
\begin{aligned}
    \mathcal L_\pi(\phi)={}&\mathcal L_Q(\phi)\\
    &+\lambda_{F,n}\Big[
    \beta_{\mathrm{corr}}\mathcal L_{\mathrm{corr}}(\phi)
    +\mathbb E_{s,a}[P_\psi(s,a)]\Big],
\end{aligned}
\label{eq:adaptive_flow_actor}
\end{equation}
where $s\sim\mathcal D_{\mathrm{on}}$,
$a\sim\pi_\phi(\cdot\mid s)$, and $\beta_{\mathrm{corr}}>0$.
Task values favor higher returns, corrections teach feasible adjustments,
and $P_\psi$ discourages actions leading to later assistance.
The critics, correction targets, and coefficients are fixed during
actor optimization; gradients pass through the generated actions.

\textbf{Relaxing corrective guidance.}
Corrective supervision helps the policy acquire feasible behavior, but
its influence should diminish as the policy learns to operate with less
assistance. We therefore reuse the intervention-to-go estimate in
\eqref{eq:filter_dependence_value} to regulate guidance strength.
At round $n$, we evaluate $P_\psi(s,a)$ on states from the latest
online batch and raw actions $a\sim\pi_\phi(\cdot\mid s)$ sampled
from the current policy. The average predicted dependence is smoothed
and used to update the guidance weight:
\begin{equation}
\begin{aligned}
    \bar P_n
    &=(1-\beta_P)\bar P_{n-1}
      +\beta_P\mathbb E_{s,a}[P_\psi(s,a)],\\
    \lambda_{F,n+1}
    &=\Pi_{[0,\lambda_{\max}]}\!\left[
      \lambda_{F,n}+\eta_\lambda
      (\bar P_n-\kappa_\lambda\lambda_{F,n})\right].
\end{aligned}
\label{eq:adaptive_filter_weight}
\end{equation}

The operator $\Pi$ denotes interval projection, with $\beta_P\in(0,1]$,
$\eta_\lambda,\kappa_\lambda>0$, and
$0<\eta_\lambda\kappa_\lambda<1$.
The feedback statistic is computed without gradient propagation, and
$\lambda_{F,n+1}$ is held fixed during the subsequent actor updates.
Predicted intervention demand counteracts weight decay; vanishing
feedback gives $\lambda_{F,n}\to0$.
Together with the decaying calibration weight $\beta_{\mathrm{cal},n}$,
this reduces the influence of recovery supervision and value
regularization, allowing continued task improvement under unchanged
execution constraints.

\subsection{Safety and Asymptotic Guidance Analysis}
\label{subsec:analysis}

SOAR-Flow separates execution protection from learning guidance.
The following result characterizes sampling-time safety and the
asymptotic removal of auxiliary contributions to policy and value updates.

\begin{theorem}
\label{thm:safety_guidance}
\textit{Certified Safety and Vanishing Auxiliary Guidance.}
Starting from a feasible $\xi_0\in\mathcal C$, suppose CBF--MPC
remains feasible and the realized transitions satisfy
$h_j(\xi_{t+1})\ge\rho_jh_j(\xi_t)-\epsilon_{j,t}$,
with $0\le\rho_j<1$ and
$0\le\epsilon_{j,t}\le\bar\epsilon_j$.
Then
\begin{equation}
    h_j(\xi_t)\ge\rho_j^t h_j(\xi_0)
    -\frac{1-\rho_j^t}{1-\rho_j}\bar\epsilon_j.
    \label{eq:closed_loop_barrier_bound}
\end{equation}

In particular, zero residual bounds preserve nonnegative barriers at
every sampling instant throughout online learning.

Let the guidance weight follow \eqref{eq:adaptive_filter_weight},
with $\bar P_0\in[0,1]$,
$\lambda_{F,0}\in[0,\lambda_{\max}]$, and
$0<\eta_\lambda\kappa_\lambda<1$.
Suppose $\bar P_n\to0$, $\beta_{\mathrm{cal},n}\to0$, and the
actor gradients of $\mathcal L_{\mathrm{corr}}$ and
$\mathbb E_{s,a}[P_\psi(s,a)]$, as well as the critic-mean gradients
$\nabla_{\theta_i}Q_{\theta_i}(s,a)$ at all calibration samples,
are uniformly bounded along the learning iterates.
Then $\lambda_{F,n}\to0$ and
\begin{equation}
\begin{aligned}
    \|\nabla_\phi\mathcal L_\pi
        -\nabla_\phi\mathcal L_Q\|&\to0,\\
    \|\beta_{\mathrm{cal},n}\nabla_{\theta_i}
        \mathcal R_{\mathrm{GCal},i}\|&\to0.
\end{aligned}
\label{eq:vanishing_auxiliary_gradients}
\end{equation}

All gradients follow the fixed-target, fixed-coefficient, and
fixed-calibration-scale conventions specified above.
\end{theorem}

The barrier bound follows by iterating the realized CBF inequality.
The guidance recursion has contraction factor
$1-\eta_\lambda\kappa_\lambda\in(0,1)$, while
$\partial\mathcal G_i/\partial Q_i=\Phi(\nu_i)\in[0,1]$
bounds the calibration gradient under the stated assumptions.
Thus, the direct influence of corrective, dependence, and calibration
terms vanishes as predicted dependence and calibration weights tend to zero.
The task and entropy objective remains active without a persistent
gradient penalty for departing from the guardian's particular recovery
choices, preserving scope for higher-return behavior within the
certified region while execution constraints remain enforced.

%% file: sections/framework_placeholder.tex
\begin{figure*}[!t]
 \centering
 \includegraphics[width=\textwidth]
 {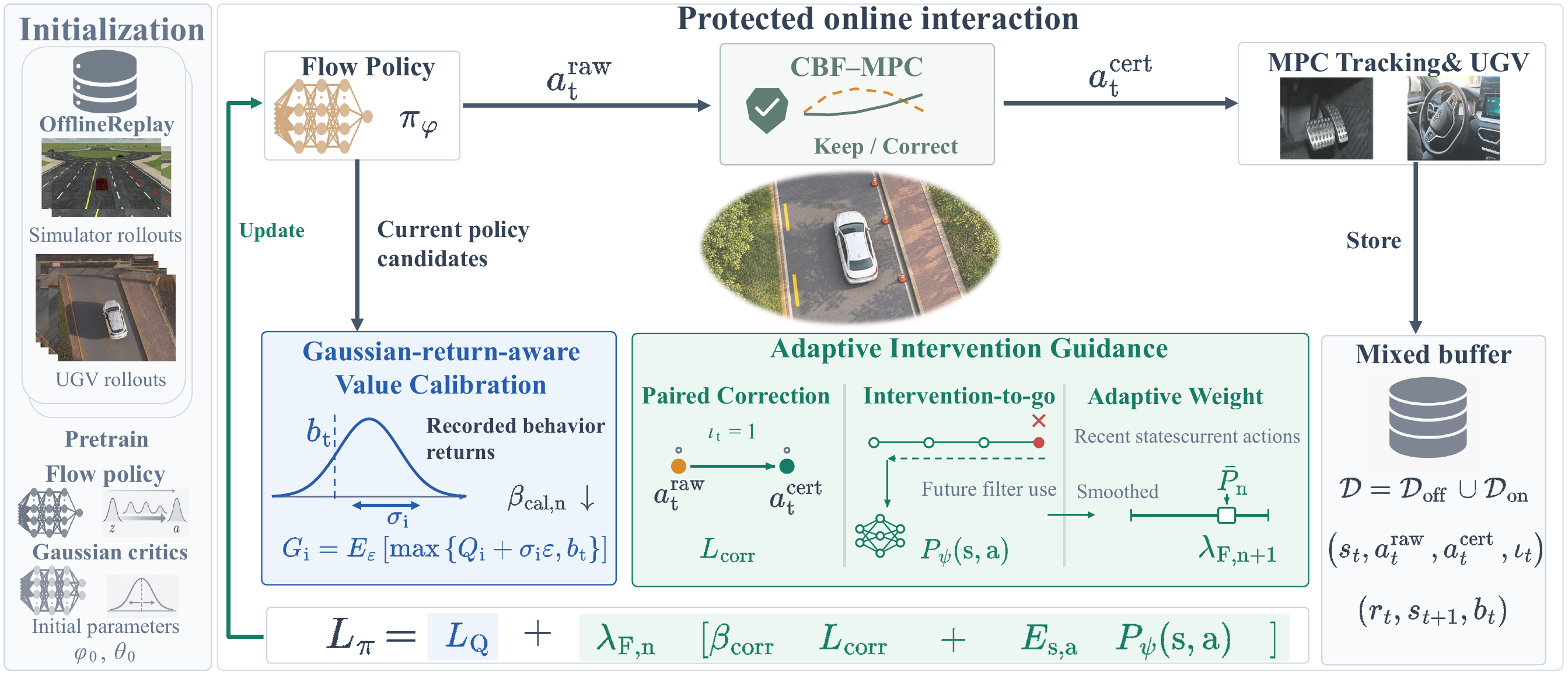}
\caption{Overall framework of SOAR-Flow.}
 \label{fig:framework}
\end{figure*}

%% file: sections/simulation_experiments.tex
\section{Simulation Experiments}
\label{sec:sim_results}

We evaluate whether SOAR-Flow improves an initialized driving policy
while limiting training violations and reducing reliance on safety
corrections across different simulated scenarios. Baseline comparisons
and ablations examine the contributions of value calibration and
adaptive guidance.

\input{sections/simulation_floats}

\subsection{Experimental Setup}
\label{sec:sim_setup}

We conduct experiments in \texttt{SafeMetaDriveEnv} of
MetaDrive~\cite{metadrive}, where scenario seeds specify different road
layouts and obstacle configurations. Fig.~\ref{fig:simulation_scenarios}
shows a scenario and the experimental workstation. An MPC controller collects offline
trajectories on ten fixed scenario seeds to initialize the Flow policy
and Gaussian task critics. Twenty additional seeds, excluded from offline
training, are used for online fine-tuning and evaluation.
The actor receives a 259-dimensional observation comprising 19 ego-state
and navigation variables, including speed, heading, steering state,
road-boundary distances, and upcoming checkpoints, together with 240
light detection and ranging (LiDAR)-like range measurements. Four
additional safety-context variables are provided only to the task critics.
The normalized actor output
$a_t=[a_t^{\mathrm{lon}},a_t^{\mathrm{lat}}]\in[-1,1]^2$
is decoded into a local target in the ego frame:
\begin{equation}
    x_t^{\mathrm{ref}}=\tfrac12(a_t^{\mathrm{lon}}+1)L,
    \qquad y_t^{\mathrm{ref}}=a_t^{\mathrm{lat}}W,
    \label{eq:exp_target_decode}
\end{equation}
with $L=10$~m and $W=3.7$~m. CBF--MPC certifies or corrects the
reference before a nominal nonlinear model predictive control (NMPC)
tracker produces steering and longitudinal commands. Task reward and safety
cost are recorded separately. The maximum commanded speed is
$10.75$~m/s. Online training uses $10^6$ environment steps per run
and five independent training seeds.

\textbf{Comparisons and metrics.}
Baselines include fixed CBF--MPC, offline Flow-BC,
Cal-QL~\cite{calql}, PROTO-SAC~\cite{proto}, Gaussian-DSAC, and
Flow-Q-only. Ablations examine corrective learning, intervention-to-go,
and calibration. PO evaluation disables CBF correction but retains NMPC,
whereas SF evaluation retains certification. PO reports episodic return,
SS, and evaluation safety cost. SS requires arrival without collision,
road departure, or nonzero episode cost.
Protected training reports cumulative safety cost. In SF evaluation,
certification-based intervention counts and rates are defined using
$\iota_t$. Repeated trials are averaged within each training seed,
followed by the mean and sample SD across seeds.

\subsection{Results and Analysis}
\label{sec:sim_analysis}

\textbf{Online improvement.}
Fig.~\ref{fig:sim_results}(a) and Table~\ref{tab:sim_results}
present the learning curves and final comparison. The final return
listed for SOAR-Flow is $352.3\pm2.1$, compared with $309.1\pm4.8$
for Flow-BC and $343.5\pm3.0$ for Gaussian-DSAC. This value is close
to the fixed CBF--MPC reference of $350.0\pm0.6$, whereas
Flow-Q-only falls below the offline baseline to $278.2\pm12.1$.
The main contrast is improvement over offline initialization;
the fixed controller provides a performance reference rather than
a learning curve.

\textbf{Safety and dependence.}
Fig.~\ref{fig:sim_results}(b--c) and Table~\ref{tab:sim_results}
summarize training cost and correction demand. SOAR-Flow is listed
with a training cost of $0.18\pm0.10$ and a correction rate of
$0.72\pm0.12\%$, compared with $0.85\pm0.32$ and
$2.68\pm0.28\%$ for Gaussian-DSAC. The table also lists
$100\%$ SS and zero evaluation cost under PO evaluation.
These metrics distinguish protected collection from execution without
reference correction: low training cost alone does not establish
autonomous task completion.

\textbf{Ablations.}
The ablation comparison is presented in Table~\ref{tab:sim_results}.
The variants without corrective learning and without $P$ have
correction rates of $4.98\%$ and $1.98\%$, respectively,
compared with $0.72\%$ for the full method. Removing calibration
and using mean-only calibration give return entries of $338.0$
and $348.0$, respectively. Mean-only calibration narrows the return
gap, but its listed correction rate remains higher at $1.20\%$.
The full-method entry combines the highest return with the lowest
training cost among these variants. PO safe-success entries are also
lower without correction or $P$, linking the tabulated comparison
to task completion as well as correction demand.

%% file: sections/simulation_floats.tex
\begin{figure}[!t]
 \centering
 \includegraphics[width=0.48\columnwidth,height=31mm,keepaspectratio]{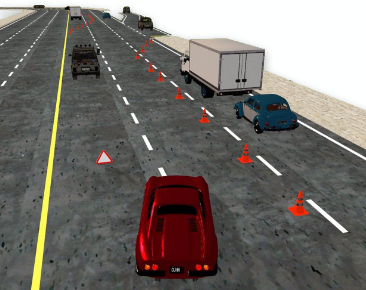}\hfill
 \reflectbox{\rotatebox[origin=c]{180}{\includegraphics[trim=0bp 0bp 0bp 1355bp,clip,width=0.48\columnwidth,height=31mm,keepaspectratio]{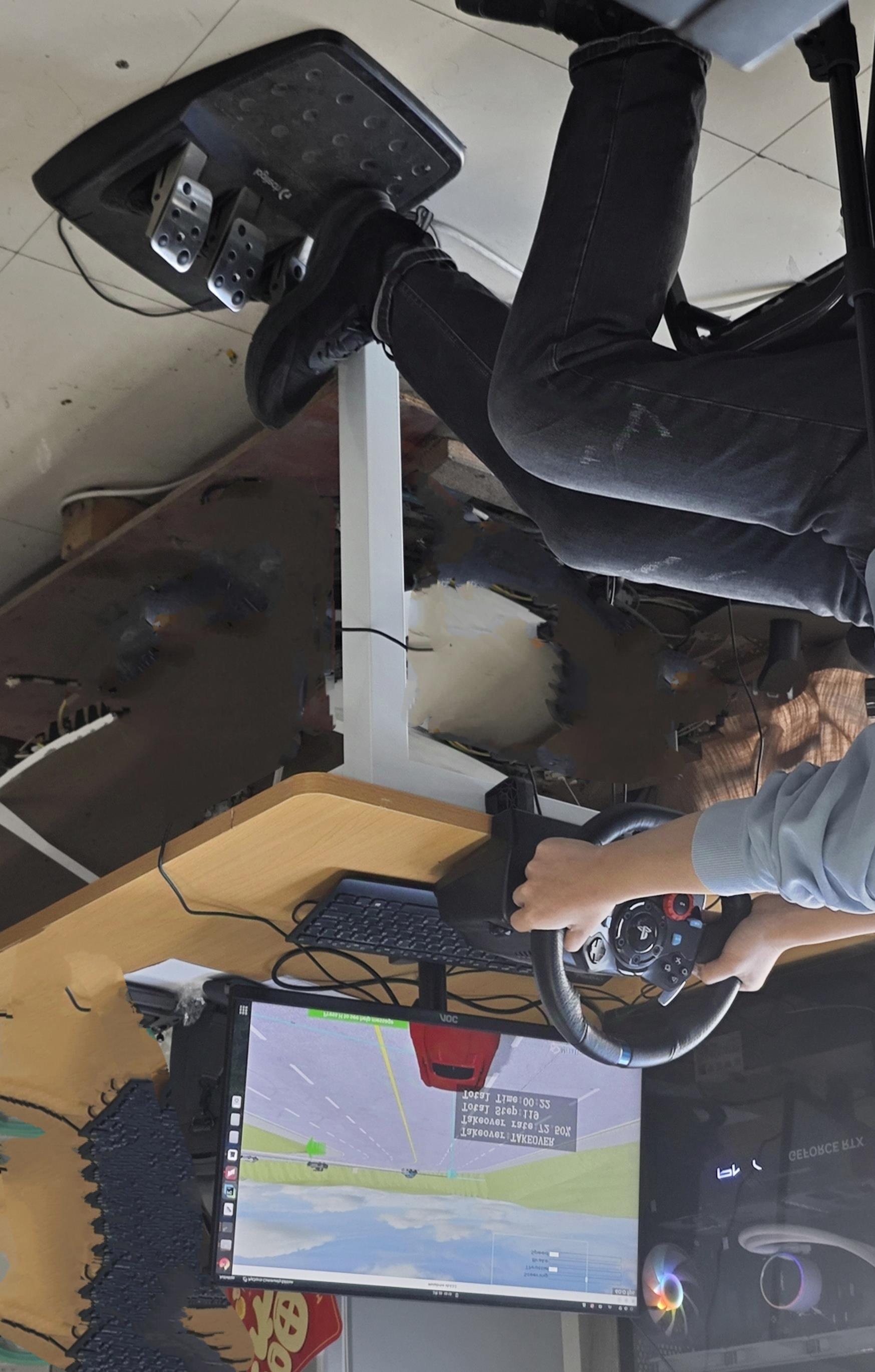}}}
 \caption{MetaDrive scenario and experimental workstation.}
 \label{fig:simulation_scenarios}
\end{figure}

\begin{figure*}[!b]
  \centering
  \captionsetup[subfloat]{captionskip=0pt}
  \subfloat[Episodic return (PO).]{%
    \includegraphics[width=0.325\textwidth]{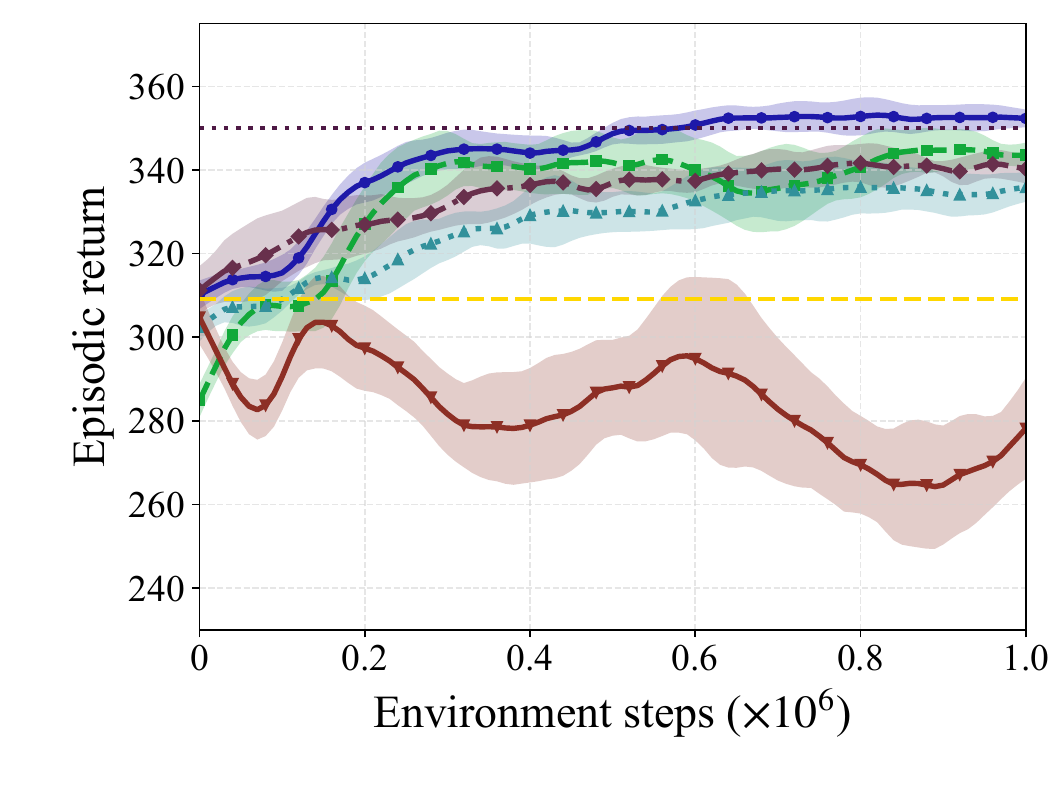}%
  }
  \hfill
  \subfloat[Training cost (SF).]{%
    \includegraphics[width=0.325\textwidth]{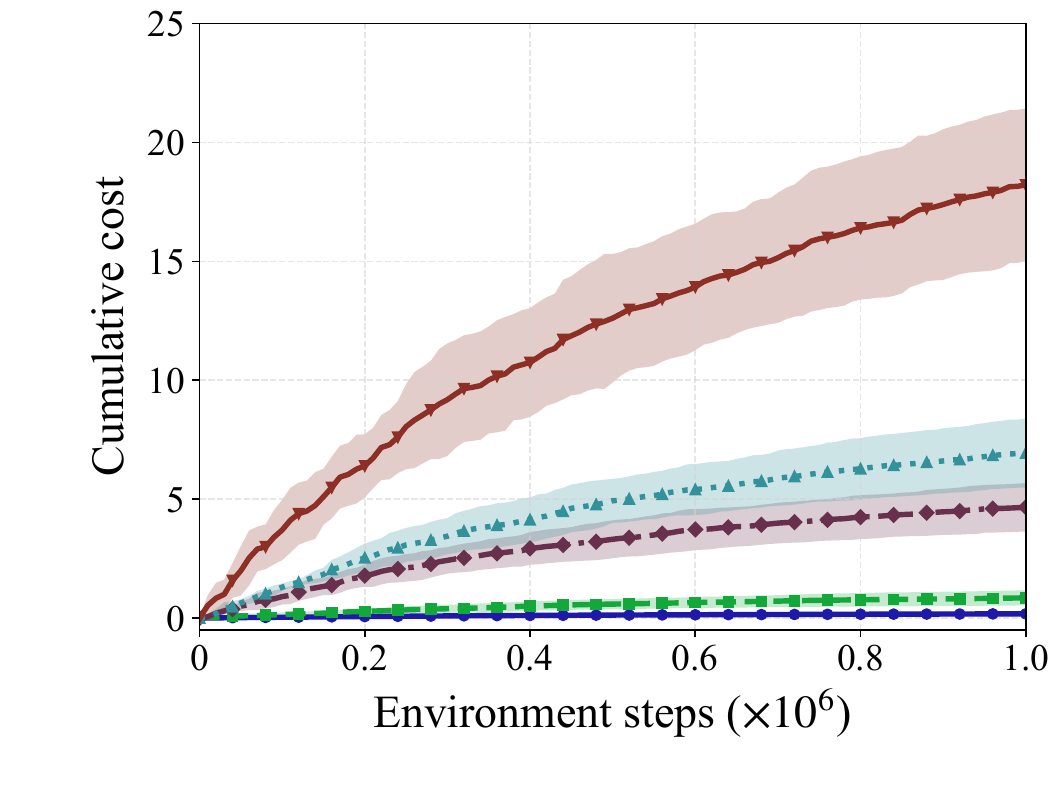}%
  }
  \hfill
  \subfloat[Correction rate (SF).]{%
    \includegraphics[width=0.325\textwidth]{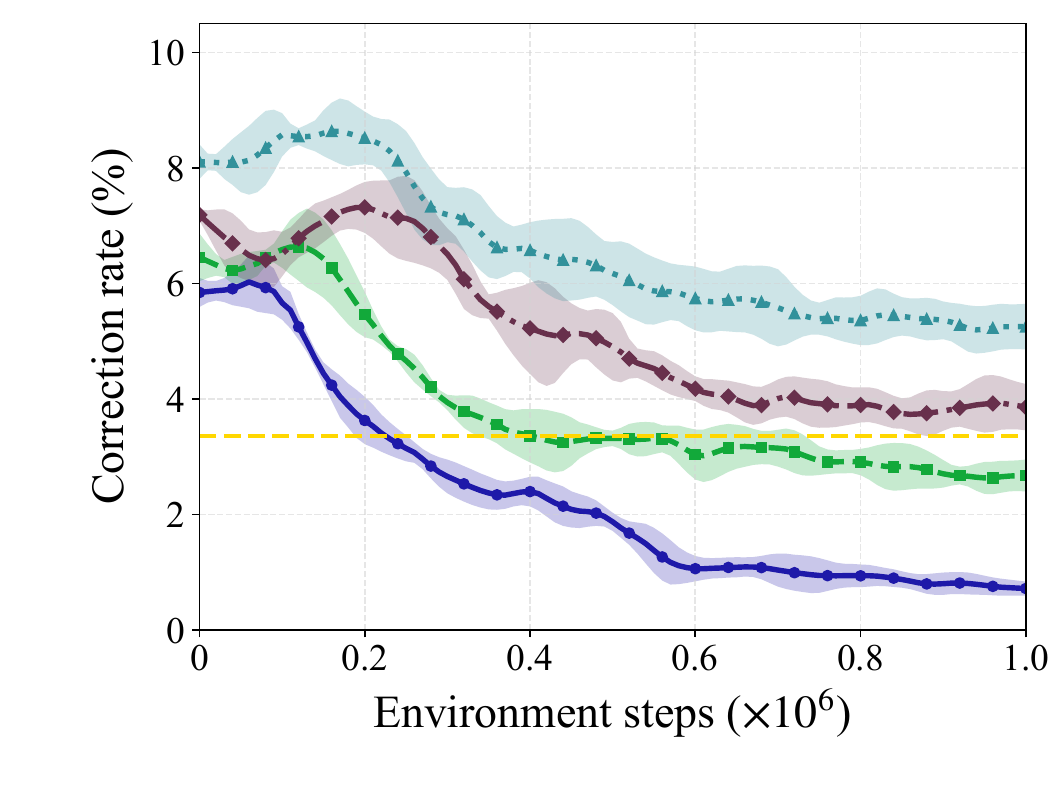}%
  }
  \par\vspace{2pt}
  \includegraphics[width=\textwidth]{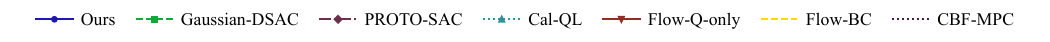}
  \caption{Simulation learning curves.}
  \label{fig:sim_results}
\end{figure*}

\begin{table*}[!t] 
  \centering
  \caption{Comparison in the simulation experiment, including safe success (SS).}
  \label{tab:sim_results}
  \small
  \setlength{\tabcolsep}{4.2pt}
  \begin{tabular}{lccccc}
    \toprule
    \textbf{Method}
      & \textbf{Return $\uparrow$}
      & \textbf{SS (\%) $\uparrow$}
      & \textbf{Eval. Cost $\downarrow$}
      & \textbf{Train Cost $\downarrow$}
      & \textbf{Corr. Rate (\%) $\downarrow$} \\
    \midrule
    CBF--MPC
      & $350.0\!\pm\!0.6$ & $\boldsymbol{100.0\!\pm\!0.0}$
      & $\boldsymbol{0.00\!\pm\!0.00}$ & -- & -- \\
    Flow-BC
      & $309.1\!\pm\!4.8$ & $70.0\!\pm\!7.1$
      & $5.42\!\pm\!0.76$ & -- & $3.33\!\pm\!0.35$ \\
    \midrule
    Cal-QL~\cite{calql}
      & $335.9\!\pm\!3.5$ & $88.0\!\pm\!3.2$
      & $1.94\!\pm\!0.54$ & $6.94\!\pm\!1.46$
      & $5.25\!\pm\!0.40$ \\
    PROTO-SAC~\cite{proto}
      & $340.3\!\pm\!3.4$ & $91.6\!\pm\!2.6$
      & $1.14\!\pm\!0.48$ & $4.66\!\pm\!1.02$
      & $3.86\!\pm\!0.40$ \\
    Gaussian-DSAC
      & $343.5\!\pm\!3.0$
      & $96.0\!\pm\!3.2$
      & $0.38\!\pm\!0.28$
      & $0.85\!\pm\!0.32$
      & $2.68\!\pm\!0.28$ \\
    Flow-Q-only
      & $278.2\!\pm\!12.1$ & $42.0\!\pm\!9.5$
      & $6.24\!\pm\!0.87$ & $18.22\!\pm\!3.20$
      & $37.11\!\pm\!4.40$ \\
    \midrule
    Ours w/o correction
      & $335.8\!\pm\!4.3$ & $86.4\!\pm\!4.8$
      & $1.82\!\pm\!0.60$ & $2.26\!\pm\!0.55$
      & $4.98\!\pm\!0.75$ \\
    Ours w/o $P$
      & $341.6\!\pm\!3.6$ & $91.6\!\pm\!3.8$
      & $0.82\!\pm\!0.35$ & $1.32\!\pm\!0.35$
      & $1.98\!\pm\!0.53$ \\
    Ours w/o calibration
      & $338.0\!\pm\!4.5$ & $92.0\!\pm\!3.2$
      & $1.20\!\pm\!0.35$ & $1.50\!\pm\!0.40$
      & $2.50\!\pm\!0.45$ \\
    Ours w/ mean calibration
      & $348.0\!\pm\!2.8$ & $98.0\!\pm\!2.0$
      & $0.20\!\pm\!0.12$ & $0.50\!\pm\!0.18$
      & $1.20\!\pm\!0.20$ \\
    Ours
      & $\boldsymbol{352.3\!\pm\!2.1}$
      & $\boldsymbol{100.0\!\pm\!0.0}$
      & $\boldsymbol{0.00\!\pm\!0.00}$
      & $\boldsymbol{0.18\!\pm\!0.10}$
      & $\boldsymbol{0.72\!\pm\!0.12}$ \\
    \bottomrule
  \end{tabular}
\end{table*}

\input{sections/ugv_comparison_table}

%% file: sections/ugv_comparison_table.tex
\begin{table*}[!b]
  \centering
  \caption{UGV comparison. SS is reported as a percentage; other metrics are mean $\pm$ SD.}
  \label{tab:real_results}
  \small
  \setlength{\tabcolsep}{2.4pt}
  \begin{tabular}{lccccccc}
    \toprule
    & \multicolumn{3}{c}{Blocked-Side Bypass}
    & \multicolumn{3}{c}{8-shaped Maneuver}
    & Held-out Cornering \\
    \cmidrule(lr){2-4}\cmidrule(lr){5-7}\cmidrule(lr){8-8}
    Method
    & SS (\%) $\uparrow$
    & Return $\uparrow$
    & Corr. Rate (\%) $\downarrow$
    & SS (\%) $\uparrow$
    & Time (s) $\downarrow$
    & Min. Clr. (m) $\uparrow$
    & SS (\%) $\uparrow$ \\
    \midrule
    CBF--MPC
    & $\mathbf{100}$
    & $345.7\!\pm\!1.5$
    & --
    & $95$
    & $16.72\!\pm\!0.31$
    & $0.779\!\pm\!0.018$
    & $85$ \\
    Flow-BC
    & $85$
    & $311.2\!\pm\!3.8$
    & $4.31\!\pm\!0.41$
    & $65$
    & $18.21\!\pm\!0.56$
    & $0.681\!\pm\!0.039$
    & $55$ \\
    Gaussian-DSAC
    & $95$
    & $343.9\!\pm\!2.6$
    & $2.46\!\pm\!0.32$
    & $90$
    & $15.96\!\pm\!0.29$
    & $0.776\!\pm\!0.017$
    & $90$ \\
    \textbf{Ours}
    & $\mathbf{100}$
    & $\boldsymbol{350.8\!\pm\!1.8}$
    & $\boldsymbol{0.68\!\pm\!0.11}$
    & $\mathbf{100}$
    & $\boldsymbol{15.36\!\pm\!0.24}$
    & $\boldsymbol{0.801\!\pm\!0.019}$
    & $\mathbf{95}$ \\
    \bottomrule
  \end{tabular}
\end{table*}

%% file: sections/ugv_experiments.tex
\section{Real-World Experiments}
\label{sec:real_results}

We examine protected online adaptation on a UGV and the development of
driving behavior requiring fewer safety corrections. A frozen-policy
test on a held-out cornering task further assesses execution beyond
the training layouts.

\input{sections/ugv_platform}

\input{sections/ugv_result_floats}

\subsection{Experimental Setup}
\label{sec:real_setup}

Our UGV in Fig.~\ref{fig:ugv_platform} is equipped with
LiDAR, cameras, radar, and an integrated navigation system (INS).
Policy inputs combine ego motion, surrounding obstacle distances, and
navigation checkpoints~\cite{h_dsac}. Policy inference and NMPC run
onboard at 10~Hz and 20~Hz, respectively. The actor specifies a local
reference; following certification, NMPC generates steering and
longitudinal commands transmitted through the base adapter unit (BAU).

Offline data are collected by a human driver in the two-sided
obstacle-bypass layout of Fig.~\ref{fig:real_results}(a).
We record 24 demonstration trajectories, with 12 passing on each side,
to pretrain the UGV Flow policy and Gaussian critics.
Online refinement alternates between \emph{Blocked-Side Bypass}
in Fig.~\ref{fig:real_results}(b), where one demonstrated passing
route is blocked, and \emph{8-shaped Maneuver}
in Fig.~\ref{fig:real_results}(c), which requires successive turns
in opposite directions around two obstacles. The first task changes
which demonstrated bypass remains feasible; the second requires the
exit from one bypass to accommodate the following turn.
A single policy is updated
across both tasks under CBF--MPC protection, with their experience stored
in the same online replay buffer. Checkpoints are evaluated on both tasks
without parameter updates to assess performance on each layout.
The final policy is frozen and tested zero-shot on
\emph{Cornering with Obstacle Avoidance} in
Fig.~\ref{fig:real_results}(d), which combines a $90^{\circ}$ bend
with two obstacles. This layout is excluded from demonstrations,
online training, tuning, and checkpoint selection. The UGV
protocol uses three independent training seeds.

\textbf{Comparisons and metrics.}
UGV comparisons use CBF--MPC, Flow-BC, and Gaussian-DSAC.
We use the PO and SF evaluation modes and SS criterion described in
Sec.~\ref{sec:sim_setup}. Each method's reported SS is computed from
20 evaluation trials per task. Learned-policy SS and return use PO;
correction rates use SF. Fixed CBF--MPC runs independently.
Table~\ref{tab:real_results} groups results by task, with continuous
metrics reported as mean $\pm$ SD.

\subsection{Results and Analysis}
\label{sec:real_analysis}

\textbf{Online adaptation.}
Fig.~\ref{fig:real_results}(e--f) and Table~\ref{tab:real_results}
show blocked-side adaptation. SOAR-Flow achieves $100\%$ SS and a return of
$350.8\pm1.8$, compared with $343.9\pm2.6$ for Gaussian-DSAC.
Its correction rate is $0.68\pm0.11\%$, versus
$2.46\pm0.32\%$ for Gaussian-DSAC and $4.31\pm0.41\%$ for Flow-BC.
The $39.6$ return gain over Flow-BC accompanies reduced correction
demand after one demonstrated route is blocked.

\textbf{Autonomous execution and efficiency.}
Fig.~\ref{fig:teaser} shows 8-shaped refinement from corrected early
and intermediate runs to late execution without CBF correction,
with nominal MPC tracking retained. The first turn must prepare for
the following opposite-direction turn. In Table~\ref{tab:real_results},
SOAR-Flow reaches $100\%$ 8-shaped SS, compared with $95\%$ for
CBF--MPC and $90\%$ for Gaussian-DSAC. Its completion time is
$15.36\pm0.24$~s, versus $16.72\pm0.31$~s for CBF--MPC, a reduction
of $1.36$~s (approximately $8.1\%$). The corresponding minimum
clearances are $0.801\pm0.019$~m and $0.779\pm0.018$~m.
Thus, shorter completion time accompanies slightly larger mean clearance.

\textbf{Zero-shot cornering.}
On held-out cornering, SOAR-Flow attains $95\%$ SS, compared with
$85\%$ for fixed CBF--MPC, $55\%$ for Flow-BC, and $90\%$ for
Gaussian-DSAC. Fig.~\ref{fig:real_results}(d) contrasts SOAR-Flow
clearing both obstacles with fixed CBF--MPC contacting the downstream
obstacle, highlighting the need to preserve clearance for subsequent
avoidance. The frozen policy transfers without further updates, although
not every evaluation succeeds.

%% file: sections/ugv_platform.tex
\begin{figure}[!htb]
  \centering
  \includegraphics[width=0.64\columnwidth]{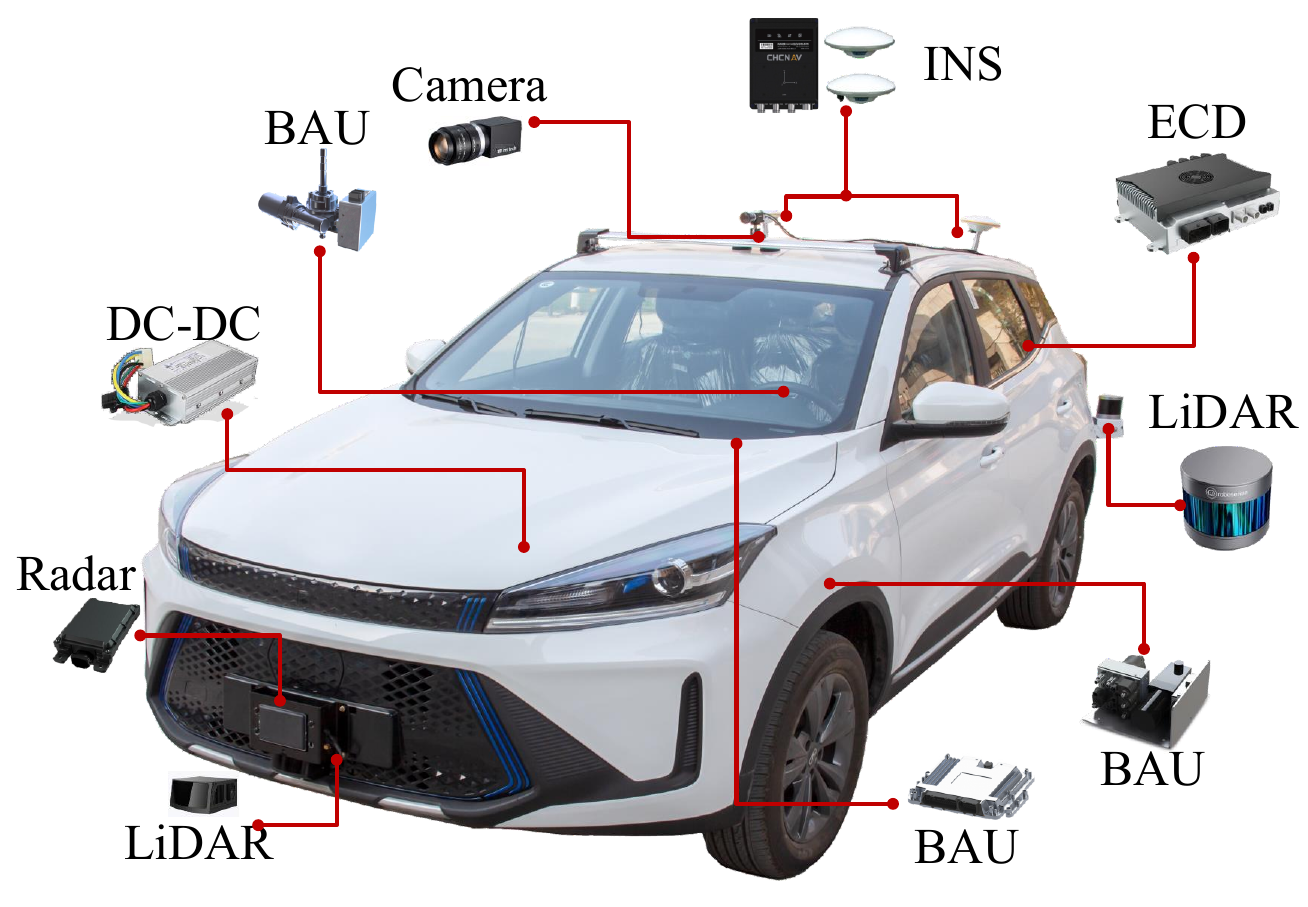}
  \caption{UGV experimental platform.}
  \label{fig:ugv_platform}
\end{figure}

%% file: sections/ugv_result_floats.tex
\begin{figure*}[!t]
  \centering
  \includegraphics[width=0.92\textwidth]{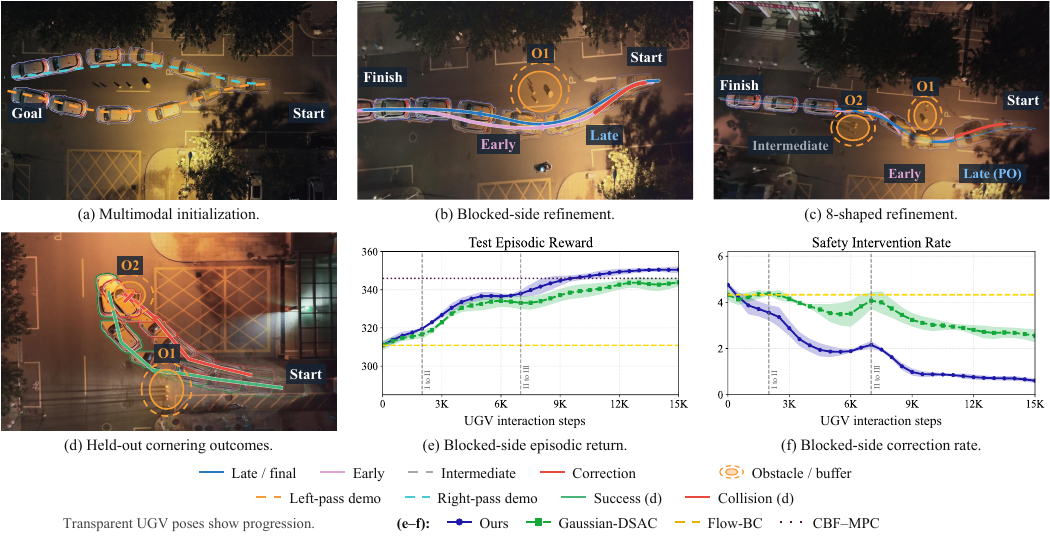}
  \caption{UGV refinement with SOAR-Flow.
  (a) Multimodal initialization. (b--c) Recorded blocked-side and 8-shaped
  trajectories at different refinement stages. Red segments indicate
  corrections; the late 8-shaped run uses PO execution with MPC
  tracking retained. (d) Held-out SOAR-Flow success and fixed CBF--MPC collision.
  (e--f) Blocked-side curves;
  vertical dashed lines mark training stages.}
  \label{fig:real_results}
\end{figure*}

%% file: sections/conclusion.tex
\section{Conclusion}
\label{sec:conclusion}

This paper presented SOAR-Flow for safe online refinement of
offline-initialized Flow-based driving policies under CBF--MPC protection.
Gaussian-return-aware calibration uses recorded behavioral outcomes
to regulate critic updates, while paired corrections and
intervention-to-go learning guide policy improvement.
Intervention-responsive guidance and decaying calibration weights
reduce auxiliary influences under the stated conditions.
The analysis establishes conditional sample-time safety of protected
execution and vanishing auxiliary gradients.
Together, these components connect protected data collection with
continued task optimization, aiming to reduce reliance on particular
recovery behaviors while retaining the prescribed execution constraints.
\input{sections/teaser}

%% file: sections/teaser.tex
\begin{figure*}[]
  \centering
  \includegraphics[width=0.92\textwidth]{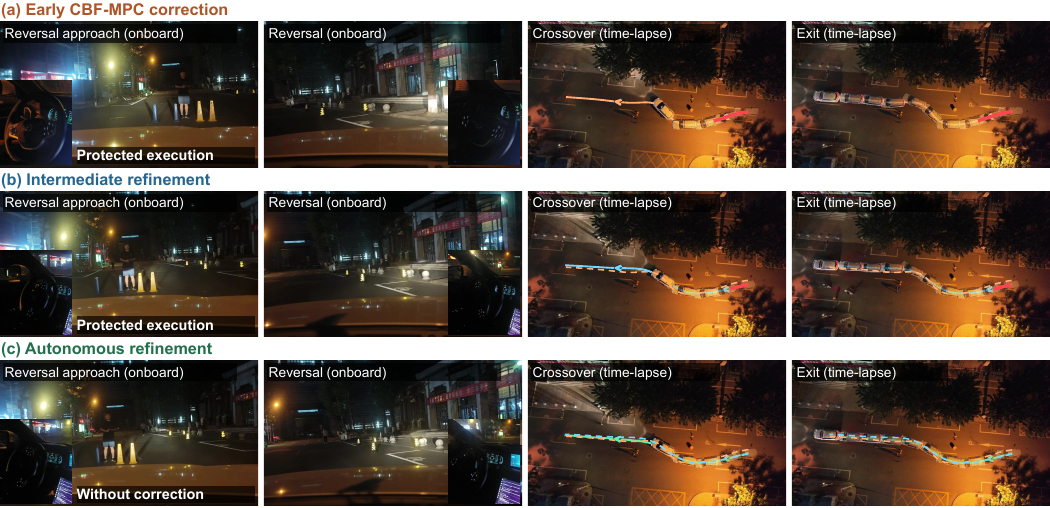}
  \caption{8-shaped refinement with SOAR-Flow: early protected execution,
  intermediate refinement, and late execution without CBF correction.
 Red segments mark correction intervals in the early
  and intermediate runs.}
  \label{fig:teaser}
\end{figure*}